\documentclass[11pt]{article}

\usepackage{amsmath}
\usepackage{url}

\usepackage{graphicx}
\usepackage{subfigure}
\usepackage{multirow}
\usepackage{colortbl}
\usepackage[letterpaper, margin=2.5cm]{geometry}

\usepackage[authoryear,sort,square]{natbib}

 \usepackage{indentfirst}

\usepackage{longtable}
\usepackage{rotating}
\usepackage{xcolor}

\usepackage{algorithm}
\usepackage{algcompatible}
\usepackage[noend]{algpseudocode}

\usepackage{amssymb}

\usepackage{multicol}
\usepackage{multirow}
\usepackage{enumitem}

\begin{document}
%
\title{10Sent: A Stable Sentiment Analysis Method Based on the Combination of Off-The-Shelf Approaches }

\author{Philipe F. Melo (1), Daniel H. Dalip(2), Manoel M. Junior(1), \\Marcos A. Gon\c{c}alves(1), Fabr\'icio Benevenuto(1) \\  (1)Federal University of Minas Gerais, Brazil \\ (2)Centro Federal de Educa\c{c}\~ao Tecnol\'ogica de Minas Gerais, Brazil \\}

\maketitle

\begin{abstract}
    Sentiment analysis has become a very important tool for analysis of social media data. There are several methods developed for this research field, many of them working very differently from each other, covering distinct aspects of the problem and disparate strategies. Despite the large number of existent techniques, there is no single one which fits well in all cases or for all data sources. Supervised approaches may be able to adapt to specific situations but they require manually labeled training, which is very cumbersome and expensive to acquire, mainly for a new application. In this context, in here, we propose to combine several very popular and effective \textit{state-of-the-practice} sentiment analysis methods, by means of an unsupervised bootstrapped strategy for polarity classification. One of our main goals is to reduce the large variability (lack of stability) of the unsupervised methods across different domains (datasets). Our solution was thoroughly tested considering thirteen different datasets in several domains such as opinions, comments, and social media. The experimental results demonstrate that our combined method (aka, 10SENT) improves the effectiveness of the classification task, but more importantly, it solves a key problem in the field. It is consistently among the best methods in many data types, meaning that it can produce the best (or close to best) results in almost all considered contexts, without any additional costs (e.g., manual labeling). Our self-learning approach is also very independent of the base methods, which means that it is highly extensible to incorporate any new additional method that can be envisioned in the future. Finally, we also investigate  a transfer learning approach for sentiment analysis as a means to gather additional (unsupervised) information for the proposed approach and we show the potential of this technique to improve our results.
\end{abstract}



\section{Introduction}

Online social media systems are places where people talk about everything, sharing their take or their opinions about noteworthy events. Not surprisingly, sentiment analysis has become an extremely popular tool in several analytic domains, but especially on social media data. The number of possible applications for sentiment analysis in this specific domain is growing fast. Many of them rely on monitoring what people think or talk about places, companies, brands, celebrities or politicians~\citep{Hu:2004:MSC:1014052.1014073,conf/epia/OliveiraCA13,DBLP:journals/corr/abs-1010-3003}.

Due to the enormous interest and applicability, many  methods have been proposed in the last few years (e.g., SentiStrength~\citep{thelwall2013heart}, VADER~\citep{hutto2014vader}, Umigon\citep{levallois2013umigon}, SO-CAL\citep{taboada2011lexicon}). In common, these methods are unsupervised\footnote{They do not require explicit manually labeling data to be used in different domains.} tools and have been applied to identify sentiments (i.e. positive, negative, and neutral) of short pieces of text such as tweets, in which the subject discussed in the text is known \textit{a priori}.
The importance of being unsupervised is that, in a real application of sentiment analysis, it can be very hard to get previous labeled data to train a classifier.

These tools are all currently acceptable by the research community as the state-of-the-art is not well established yet. However, a recent effort~\citep{ribeiro2015benchmark} has shown that the prediction performance of these methods varies considerably from one dataset to another. For instance, in that study, Umigon was ranked in the first position in five datasets containing tweets and was among the worst in a dataset of news comments. Even among similar datasets, existing methods showed \textbf{low stability}  in terms of their ranked positions.
This suggests that existing unsupervised approaches should be used very carefully, especially for unknown datasets. More importantly, it suggests that novel sentiment analysis methods should not only be superior to existing ones in terms of predictive performance, but they should also be stable, that is, its relative prediction performance should vary minimally when used in many different datasets and contexts.

Accordingly, in this article, we propose 10SENT, an unsupervised learning approach for sentence-level sentiment classification that tells if a given piece of text (i.e. a tweet) is positive, negative, or neutral. In order to obtain better results than existing methods and guarantee stability across datasets, our approach exploits the  combination of their  classification outputs in a smart way. Our strategy relies on using a bootstrapped learning classifier that creates a training set based on a combination of answers provided by existing unsupervised methods. The intuition is that if the majority of the methods label an instance as positive, it is likely that it is positive, and it could be used to learn a classifier. This self-learning step provides to our method a level of adaptability to the current (textual) context, reducing prediction performance instability, a key aspect of an unsupervised approach.

We test our proposed approach by combining the top (best) ranked methods, according to a recent benchmark study~\citep{ribeiro2015benchmark}. We evaluate 10SENT with thirteen gold standard datasets containing social media data from different sources and contexts. Those datasets consist of different sets of labeled data annotated for positive, negative and neutral texts from social networks messages and from comments of news articles, videos, websites, and blogs.  Our approach showed to be statistically superior to (or at least ties with) the existing individual methods in most datasets. As a consequence, our approach obtains the best mean rank position considering all datasets. Thus, our experimental results demonstrate that our combined method not only improves significantly the overall effectiveness in many datasets but its cross-dataset performance variability is minimal (maximum stability). In practical terms, this means that one can use our approach in any situation in which the base methods can be exploited, without any extra cost (since it is unsupervised) and without the need to discover the best method for a given context, and still obtain top-notch effectiveness in most situations.

We also show that 10SENT is superior to basic baseline combinations, such as a majority voting approaches, with gains of up to 17\% against such baselines. This highlights the importance of our bootstrapped strategy to improve the effectiveness of the sentiment classification task. It is important to stress that the number of methods to be combined is not necessarily restricted to ten. Our self-learning approach is very independent of the base methods, which means that it is highly extensible to incorporate any new additional method that can be created in the future.

To summarize, the main contribution of our work is an easily deployable and stable method that can produce results as good or better than  the best single method for most datasets (the performance of the base methods can vary a lot) in a completely unsupervised manner, being much superior than other unsupervised solutions such as majority voting and, in some cases, close to the best supervised ones. As far as we know, this is the first time non-trivial unsupervised learning is used along with ``state-of-the-practice'' sentiment analysis methods to solve important issues in the field such as stability, generality, and improved effectiveness, all at the same time.

Finally, as a second contribution, we start an investigation into a important question of our research: whether we can ``transfer'' some knowledge to our method from a dataset labeled with emoticons by Twitter users, which is easily available, meaning that no extra labeling effort is necessary. The main idea here is that such transfer of knowledge could provide additional (unsupervised) information to our method helping to improve it even further. 


\section{Related Work}




There are currently two distinct categories of sentiment analysis methods used in the social media domain: lexicon-based and those based on machine learning techniques. Machine learning methods comprise supervised classifiers trained with labeled datasets in which classes correspond to polarities (e.g. positive, negative or neutral) \citep{pang2002thumbs}.One major challenge in this scenario is the difficulty in obtaining  annotated data to train (supervised) methods due to issues such as cost and the inherent complexity of the labeling task. Accordingly, in here, we propose an unsupervised solution to deal with this sentiment analysis task.

Lexicon-based methods exploit lexical dictionaries, that is, word lists associated with sentiments or other specific features, which are usually not based on supervised learning. some challenges with lexicon-based solutions including the construction of  the lexicon itself (which is usually manually done) and difficulties in adapting for domains different from which they were originally designed.

Such issues naturally call for a combination of solutions that exploits their strengths while overcome their limitations. The idea of combining different sentiment analysis strategies, however, has been only recently explored, but most of the existing literature on the combination of sentiment analysis involves a learning component.

For instance, \cite{prabowo2009sentiment} proposed a new hybrid classification method based on the combination of  different strategies. This work combines a rule-based classification and other  supervised learning strategies into a new hybrid sentiment classifier. \cite{DangZC10} combined machine learning and semantic-orientation that consider words expressing positive or negative sentiments.

\cite{zhang} explored an entity-level sentiment analysis method specific to the Twitter data. In that work, the authors combined lexicon and learning-based methods in order to increase the recall rate of individual methods. Differently from our work, this method was proposed for the entity-level, while ours focus on a sentence-level granularity. Similarly, 
\cite{Mudinas2012} proposed \textit{pSenti}, a method for sentiment analysis developed as a combination of lexicon and learning approaches for a different granularity level, the concept-level (semantic analysis of texts by means of web ontologies or semantic networks).

\cite{MoraesVPDAG13} investigated approaches to detect the polarity of FourSquare tips using supervised (SVM, Maximum Entropy and Na{\"i}ve Bayes) and unsupervised (SentiWordNet) learning. They also investigate hybrid approaches, developed as a combination of the learning and lexical algorithms. All techniques were tested separately and combined, but the authors did not obtain significant improvements with the hybrid approaches over the best individual techniques for this particular domain.

 \cite{polly2016offtheshelf} analyzed different datasets and considered supervised machine learning in the context of classifiers' ensembles. Their methodology also consisted of combining a set of different sentiment analysis method in a ``off-the-self'' strategy to generate the ensemble method. Their results suggest that it is possible to obtain significant improvements with ensemble techniques depending on the domain. In here, we focus on a unsupervised solution enhanced with an automatic bootstrapping step.


In a more recent effort on the ensemble direction, \cite{gonccalves2013comparing} exploits the power of the combination of some of the state-of-the-art methods, showing that they can outperform individual methods. Their results show the potential of simple solutions such as  majority voting, but the authors did not delve deep in more complex combination strategies.

Some approaches use a limited amount of labeled data (also known as weakly supervised classifier) in order to predict the sentiment in some domains. For example, \cite{siddiqua2016} proposed a weakly supervised classifier for Twitter sentiment analysis. In this work,  Naive-Bayes (NB) is combined with a rule-based classifier based on several publicly available sentiment lexicons to extract positive and negative sentiment words.  After the rule-based classifier is applied, the NB is used to classify the remaining tweets as positive or negative.

\cite{deriu2017leveraging} also uses a weakly supervised approach to multi-language sentiment classification task. The developed method  evaluates large amounts of weakly supervised
data in various languages to train a multi-layer convolutional neural network, but its focus is on multilingual sentiment classification.

Wikisent, proposed by \citep{mukherjee2012wikisent} also describes a weakly supervised system for sentiment analysis classification. They use text summarization focused on movie reviews domain in order to obtain knowledge about the various technical aspects of the movie. After that, the summary of the opinions are classified by using the SentiWordNet lexicon method.


To summarize, many authors proposed supervised ensemble classifiers, but differently from those, we propose a novel approach by combining a series of "state-of-the-practice'' existing methods in a totally unsupervised and in much more elaborated manner exploiting bootstrapping and (unsupervised) transfer learning. Another major difference of our effort is that we evaluate using multiple labeled datasets, covering multiple domains and social media sources. This is critical for an unsupervised approach given that the performance of the base methods varies significantly. As we shall see, our solution produced the most consistent results across all datasets and contexts.

\section{Combining Methods}\label{sec:methods}

Sentiment analysis can be applied to different tasks. We restrict our focus on combining those efforts related to detect the polarity (i.e. positivity, negativity, neutrality) of a given short text (i.e. sentence-level).  In other words, given a set $S$ of opinionated sentences, we want to determine whether each sentence $s$ $in$ $S$ expresses a positive, negative or neutral opinion.
We focus our effort on combining only unsupervised ``off-the-shelf'' methods. Our strategy consists of using the output label predicted by each individual method as input for a bootstrapping technique -- a self-starting process supposed to proceed without external input. Next we present the proposed technique.

Our \textbf{bootstrapping technique} is an unsupervised machine learning algorithm that uses the sentiment scores produced by each individual sentiment analysis method to create a training set for a supervised machine learning algorithm. With this algorithm, we are able to produce a final result regarding the sentiment of a sentence. Note that, we did not need to use any manually labeled data in order to produce the model.

\sloppy
We describe the method in Algorithm~\ref{alg:bootstrapping}. Suppose we have access to a set of sentences $S = \{s_{tr0}, s_{tr1}, s_{tr3},...,s_{tr_n}\}$, which are candidates of being part of our training data. Our goal is to use the unlabeled data $S$ in order to produce a training set $train$ and, then, apply it to unseen sentences for which we want to predict (here represented as $test = \{s_{tst0}, s_{tst1}, s_{tst3},...,s_{tst_m}\}$), generating the set of predictions $P$.  The training data $train$ is represented by a set of pairs $(c,s)$ where $c$ is the class representing a sentiment (positive, negative or neutral) obtained by using the information of each sentiment analysis method described in Section~\ref{sec:methods} and $s$ is a sentence represented by a set of features which, in our case, corresponds to the off-the-shelf sentiment methods' outputs.

\begin{algorithm}
    \centering
    \small
    \caption{Bootstrapping Algorithm}
    \label{alg:bootstrapping}
    \begin{algorithmic}[1]
    {
      \Require Minimum of Agreement $A$
      \Require Minimum of Confidence $C$
      \Require The set of $n$ sentences $S = s_{tr0}, s_{tr1}, s_{tr3},...,s_{tr_n}$, candidates of being part of our training data
      \Require The set of $m$ sentences which we want to predict:  $test = \{s_{tst0}, s_{tst1}, s_{tst3},...,s_{tst_m,}\}$

      \State Let $train$ = our training set represented by $(c,s)$ which $c$ is the target class and $s$ is the sentence

      \State Let $P$ = our result which is represented by a set of triplet $(i, predicted\_class, confidence)$ which is the instance, the predicted class and its confidence

      \ForAll{$s \in S$ }
            \If{$agree(s) \ge A$}
                \State \textbf{Add} the pair $(agreeClass(s),s)$ to $train$
                \State \textbf{Remove} $s$ from $S$
            \EndIf
      \EndFor
      \State \textbf{Create} a model $M$ using $train$
      \State \textbf{Apply} the model $M$ in $S$ to obtain the predictions $P$

      \ForAll{$(s, predicted\_class, confidence) \in P$ }
        \If{$confidence \ge C$}
             \State \textbf{Add} the pair $(predicted\_class,s)$ to $train$
        \EndIf

      \EndFor
      \STATE \textbf{Create} a model $M$ using $train$\\

      \STATE \textbf{Apply} the model $M$ in $test$ to obtain the predictions $P$

    }
    \end{algorithmic}
\end{algorithm}

The $test$ is represented by a set of sentences $test = \{s_{tst0}, s_{tst1}, s_{tst3},...,s_{tst_n}\}$ and, the prediction $P$, contains a set of triplets  $(s, predicted\_class, confidence)$ representing the sentence, the predicted class and the confidence (i.e. a score representing how confident the machine learning method is in its prediction), respectively.

We use the function  $agree(s)$, for each sentence $s$, which computes the Agreement level, in other words, the maximum number of sentiment analysis methods agreeing with each other regarding the sentiment in the sentence $s$. If this number is higher than the threshold $A$, we add the sentence $s$ in the training set $train$, removing it from $S$. Note that, when adding a sentence to $train$ we use the method $agreeClass(s)$ in order to obtain the class $c$ which will the sentiment assigned to $s$. Class $c$ is obtained by using the class which has the majority of sentiment analysis methods assigned to the sentence $s$.

After doing this for all the sentences in $S$, only sentences for which we could not infer a label with enough agreement remain in $S$. Then, in order to increase our training data, we use our training set $train$ to train a classification model and apply it to sentences in $S$, producing the predictions $P$. By doing so, we are able to use $P$ to add more sentences to $train$. In order to avoid noise, we only add sentences for which the learned model produces a confidence higher than a threshold $C$.
Finally, we retrain with the new set $train$ and apply it to $test$ in order to produce, for each sentence $s$, a single score $c$ representing its final sentiment score.



As mentioned before, our approach consists of combining popular \textbf{``off-the-shelf'' sentiment analysis methods} freely available for use.  It is important to highlight that the number of methods to be combined is not necessarily restricted to ten. In fact, there is no  limit on the number of methods we can include as part of our approach -- thus, we focus on the ones evaluated by
\cite{ribeiro2015benchmark} as it provides the most recent and complete sentence-level benchmark of off-the-shelf sentiment analysis methods.


There are few small adaptations on some methods to provide as output positive, negative and neutral decisions. For this, we have used the codes shared by the authors of~\cite{ribeiro2015benchmark}. More details about these implementations can be found there. The considered methods include: VADER~\citep{hutto2014vader}, AFINN\citep{nielsen2011new}, OpinionLexicon~\citep{Hu:2004:MSC:1014052.1014073}, Umigon~\citep{levallois2013umigon}, SO-CAL~\citep{taboada2011lexicon}, Pattern.en~\citep{smedt2012pattern}, Sentiment140~\citep{mohammad2013nrc},  EmoLex~\citep{journals/ci/MohammadT13},
Opinion Finder\citep{wilson2005opinionfinder}, and SentiStrength~\citep{thelwall2013heart}. A  brief description of these methods can also in found in~\cite{ribeiro2015benchmark}.


We also note that all  methods exploit light-weight unsupervised approaches that rely on lexical dictionaries, usually implemented as a hash-like data structure. For this reason, the execution performance of our combined, as well as the individual methods, does not require any powerful hardware platform.

\section{Methodology}

Next we present the gold standard datasets used to evaluate our approach, the combined baseline method, the metrics used for evaluation and the experimental setup.


In our evaluation, we use 13 \textbf{datasets} of messages labeled as positive, negative and neutral from several domains, including messages from social networks, opinions and comments in news articles and videos.
These datasets were kindly shared by the authors of ~\citep{ribeiro2015benchmark}.  We only consider those with three classes (positive, negative, and neutral). The number of messages vary from few hundreds to a few thousands. The datasets are usually very skewed, with usually one or two classes outnumbered the majority one by large margins. The median of the average number of phrases pier message is around 2 while the average number of words vary from around 15 to approximately 60. We refer the reader to their work for more details about the datasets. We emphasize that the diversity and amount of different datasets used in our evaluation allow us to accurately evaluate not only the prediction performance of the proposed method, but also measure the extent to which a method's result varies when it is tested for different social media sources.

\sloppy

Regarding the \textbf{baseline}, as 10SENT explores the output of 10 other individual methods,  Majority Voting is a natural baseline\footnote{Notice that Weighted Majority Voting is not an option as a fair baseline, since to determine the weights we would need some type of supervision, something that our method does not exploit. In any case, we compare our solution to a version of the Weighted Majority Voting method in the `Upperbound Comparison' Section.}. Voting is one of the simplest ways to combine several methods. By assuming that each individual method gives us a unique label as output for a sentence, the final result of Majority Voting is the label that the majority of the base classifiers returned as output for that sentence\footnote{In this method, ties are possible. In this case, we assign a \textit{Neutral} class to the sentence.}.

The major advantages of this approach are its simplicity and extensibility, i.e., it is very easy to include new (off-the-shelf) methods. Also, no training data is necessary  for this method, which fits well with our purpose of an unsupervised solution. On the other hand, majority voting  is not as flexible as 10SENT in coping with  all the diversity of the methods.
This is due to the training phase of 10SENT that allows it to capture some idiosyncracies of each one of them.

As \textbf{evaluation metric} the use the popular Micro and Macro-F1 scores. Micro-F1 captures the overall accuracy across all classes. Macro-F1 calculates the F1 score for each class separately report the average of these scores for all classes. It is important in datasets with high skewness (as is the case here) or in problem in which we are more interested in the effectiveness in the majority classes.

As a third evaluation metric, we use \textit{Mean Ranking}. As we have a potential large number of results, considering all base methods and
datasets, it is important to have a global measure of performance for all these combinations in a single metric. For doing so, we ranked the methods for each dataset.
The Mean Ranking is the simple sum of ranks obtained by a method in each dataset divided by the total number of datasets. It is
important to notice that the rank was calculated based on Macro F1 because of the high imbalance among the classes in several datasets.

Finally, our \textbf{experiments} were run using a 5-fold cross validation setup, with best parameters for the learning methods found using cross-validation within the training set. This procedure was applied to all considered datasets. To compare the average results in the test sets of our  experiments, we assess the statistical significance of our results by means of a paired t-test with 95\% confidence. We just consider statistically significant, results whose the value of $p$ is less than 0.05 and any stated claim of superiority is based on these tests. Finally, we adapted the original outputs values of base methods to our corresponding polarities. In particular, an output equals to zero was considered as a neutral or ``absence of opinion''.



\section{Experimental Results}

Here, we discuss some decisions taken during the development of 10SENT and start some investigation on issues related to transfer learning for sentiment analysis, showing the potential of this technique to improve our results.

\subsection{Choice of The Classifier}

10SENT is an unsupervised machine learning method as it does not exploit manually labeled data, only the agreement among the base methods. Given that the bootstrapping process adds a set of instances with high confidence into a training set, it is possible to perform a learning step exploiting  such data in the the usual format training/validation. Because of this, there is a need to investigate which classifier fits better this application. Thus, we perform a series of tests with our method using different classification algorithms in order to choose the best  one for this task. In all these tests, we used all 10 methods of 10SENT. We tested three different and widely used algorithms in our approach:  \textit{Support Vector Machine} (SVM) \citep{ChangSVM}, \textit{Random Forest} (RF) \citep{randomforest} and\textit{ $k$-Nearest Neighbors} (KNN) \citep{knn}. Here we used the implementations of RF and KNN provided in scikit-learn\footnote{Available at \url{http://scikit-learn.org/stable/index.html}} and for SVM, we use LibSVM\footnote{Available at \url{http://www.csie.ntu.edu.tw/~cjlin/libsvm/}} package. Specifically, we use a \textit{radial basis function} (RBF) kernel with a grid search for the best parameters. Overall, Random Forests produced the best results in most datasets, being the final choice for our bootstrapping method.

\subsection{Choice of Number of Methods}

To verify the coherence with results obtained by Majority Voting, we perform a test with different number of methods used in the combination. In this test, we want to check how the addition of a method can impact the outcome. 
We evaluated results of 10SENT combining from 3 up to 10 methods. 
In these experiments, we included from the best to the worst method in each dataset, according to~\cite{ribeiro2015benchmark}. We noted that adding a new method improves the overall results, but it is possible to note that improvements get smaller with new inclusions. Thus, after a certain number, the gain is minimal. Therefore, we fixed 10 as a good choice to number of methods in 10SENT core.

\subsection{Choice of Parameters}					
	
In our method, we need to define two important parameters: the agreement and the confidence level. Accordingly, we performed a study to better understand how our method performs when varying such parameters.
In more details, the first tested parameter was the minimum number of agreements among the methods we should use in the first round of classification (Agreement Level).


Table \ref{table:concordance} shows Macro-F1 results for each number of agreements. As we have a total of ten base methods, this table shows bootstrapping results when we use instances that at least 4 or more methods agree with each other, 5 and so on. We did not show results with less than 3 agreements since there were no instances in such scenario.

\begin{table}[!htpb]
    \centering
    \small
    \begin{tabular}{|l|l|l|l|l|l|l|l|l|}
    \hline
        \multicolumn{9}{|c|}{\textbf{\#Concordants}}                                                                                             \\ \hline
        \textbf{DATASET}                & \textbf{3} & \textbf{4} & \textbf{5} & \textbf{6} & \textbf{7} & \textbf{8} & \textbf{9} & \textbf{10} \\ \hline
        \textbf{english\_dailabor}      & 69.58      & 69.18      & 69.23      & 68.50      & 66.91      & 64.81      & 59.58      & 60.75       \\ \hline
        \textbf{aisopos\_ntua}          & 60.93      & 60.54      & 56.87      & 57.74      & 64.14      & 59.65      & 54.58      & 58.93       \\ \hline
        \textbf{tweet\_semevaltest}     & 63.54      & 63.58      & 63.64      & 63.47      & 63.82      & 61.56      & 59.36      & 59.15       \\ \hline
        \textbf{sentistrength\_twitter} & 56.75      & 58.32      & 59.13      & 58.34      & 57.58      & 55.35      & 54.27      & 57.44       \\ \hline
        \textbf{sentistrength\_youtube} & 55.63      & 55.22      & 55.67      & 56.39      & 56.65      & 55.44      & 54.69      & 54.50       \\ \hline
        \textbf{sanders}                & 56.23      & 55.72      & 55.94      & 55.64      & 55.27      & 52.73      & 50.77      & 48.14       \\ \hline
        \textbf{sentistrength\_digg}    & 51.13      & 51.29      & 53.86      & 54.58      & 51.51      & 50.98      & 48.06      & 51.83       \\ \hline
        \textbf{sentistrength\_myspace} & 46.76      & 48.30      & 48.71      & 50.31      & 50.52      & 51.02      & 54.34      & 39.44       \\ \hline
        \textbf{sentistrength\_rw}      & 48.96      & 50.09      & 46.62      & 49.73      & 49.00      & 46.52      & 46.03      & 47.58       \\ \hline
        \textbf{sentistrength\_bbc}     & 49.15      & 50.62      & 46.95      & 47.41      & 46.31      & 45.04      & 45.75      & 46.57       \\ \hline
        \textbf{debate}                 & 45.61      & 45.82      & 43.69      & 43.97      & 44.47      & 44.99      & 43.57      & 43.10       \\ \hline
        \textbf{nikolaos\_ted}          & 46.29      & 44.71      & 48.13      & 46.43      & 46.97      & 47.52      & 44.50      & 47.24       \\ \hline
        \textbf{vader\_nyt}             & 36.13      & 36.19      & 36.32      & 37.35      & 38.21      & 36.89      & 32.54      & 34.02       \\ \hline
    \end{tabular}
    \caption{Comparative table of results (F1) for 10SENT bootstrapping by different agreement levels among the base methods in classification}
    \label{table:concordance}
\end{table}

As we can see in this table, the extreme cases of agreement or disagreement produce the worst results. There is a small amount of instances with 100\% of agreement, which harms the training of the algorithm.
On the other hand, when the agreement is very low, there is a lot of noise in the training data. In sum, the Agreement level represents a trade off between the amount of available data for training and the amount of noise.
	
The second parameter was the RF confidence Level, defined in Algorithm~\ref{alg:bootstrapping} as the constant $C$. The Confidence Level is the confidence ratio of the Random Forest algorithm in its predictions. We use this in order to add more data to train during the bootstrapping step.
Then, a similar variation of this parameter was tested, as shown in Table \ref{table:confidence}.

\begin{table}[!htpb]
    \centering
    \small
    \begin{tabular}{l|l|l|l|l|l|l|l|l|}
    \cline{2-9}
                                                 & \multicolumn{8}{c|}{Confidence Level}                                                                                                                                                                                 \\ \hline
    \multicolumn{1}{|c|}{DATASET}                & \multicolumn{1}{c|}{0.3} & \multicolumn{1}{c|}{0.4} & \multicolumn{1}{c|}{0.5} & \multicolumn{1}{c|}{0.6} & \multicolumn{1}{c|}{0.7} & \multicolumn{1}{c|}{0.8} & \multicolumn{1}{c|}{0.9} & \multicolumn{1}{c|}{1.0} \\ \hline
    \multicolumn{1}{|l|}{english\_dailabor}      & 67.80                    & 66.82                    & 67.28                    & 67.50                    & 67.65                    & 67.63                    & 67.57                    & 67.64                    \\ \hline
    \multicolumn{1}{|l|}{aisopos\_ntua}          & 64.53                    & 64.19                    & 63.81                    & 64.95                    & 66.21                    & 60.89                    & 57.57                    & 57.36                    \\ \hline
    \multicolumn{1}{|l|}{tweet\_semevaltest}     & 62.56                    & 62.88                    & 62.75                    & 62.65                    & 62.82                    & 63.33                    & 63.21                    & 63.02                    \\ \hline
    \multicolumn{1}{|l|}{sentistrength\_twitter} & 58.11                    & 58.08                    & 58.47                    & 59.24                    & 58.14                    & 56.75                    & 57.71                    & 56.12                    \\ \hline
    \multicolumn{1}{|l|}{sentistrength\_youtube} & 56.64                    & 56.50                    & 55.59                    & 55.55                    & 56.28                    & 56.93                    & 56.86                    & 56.04                    \\ \hline
    \multicolumn{1}{|l|}{sanders}                & 55.22                    & 55.88                    & 54.83                    & 54.62                    & 55.65                    & 54.70                    & 53.81                    & 53.78                    \\ \hline
    \multicolumn{1}{|l|}{sentistrength\_myspace} & 51.86                    & 51.77                    & 52.89                    & 52.46                    & 55.22                    & 52.75                    & 54.51                    & 53.03                    \\ \hline
    \multicolumn{1}{|l|}{sentistrength\_digg}    & 51.75                    & 50.98                    & 53.15                    & 51.93                    & 52.59                    & 51.11                    & 50.86                    & 50.85                    \\ \hline
    \multicolumn{1}{|l|}{sentistrength\_rw}      & 46.46                    & 45.60                    & 50.24                    & 49.61                    & 50.84                    & 50.59                    & 45.77                    & 47.30                    \\ \hline
    \multicolumn{1}{|l|}{sentistrength\_bbc}     & 45.67                    & 45.75                    & 46.16                    & 45.17                    & 47.19                    & 48.00                    & 46.01                    & 48.24                    \\ \hline
    \multicolumn{1}{|l|}{debate}                 & 45.77                    & 45.95                    & 46.10                    & 46.53                    & 45.48                    & 45.26                    & 43.94                    & 43.93                    \\ \hline
    \multicolumn{1}{|l|}{nikolaos\_ted}          & 45.80                    & 44.70                    & 43.05                    & 46.42                    & 44.76                    & 46.00                    & 45.79                    & 43.48                    \\ \hline
    \multicolumn{1}{|l|}{vader\_nyt}             & 38.53                    & 38.33                    & 38.49                    & 38.65                    & 37.69                    & 37.27                    & 36.94                    & 36.10                    \\ \hline
    \end{tabular}
	\caption{Comparative table of results (F1) for 10SENT by different confidence levels added to training in classification.}
	\label{table:confidence}
\end{table}

As a final conclusion of these experiments, we arrive at a value of 7 for agreement and 0.7 for confidence, in most datasets, as a way to achieve the ``best'' balance between quantity and quality for the training data.

\subsection{Bag of Words vs. Predictions}

After the definitions of the parameters for the classification process, additional features can be extracted and combined with the predictions of other methods to improve results. One example is the text of messages itself. With the text, we can extract the Bag of Words representation (BoW) of the sentences included in the training. We used the traditional TF-IDF representation calculated for each sentence in each dataset. This was concatenated with the results of each method, as in the traditional 10SENT.

In Table \ref{tab:bow} we find the results comparing the use of these different sets of features, the predictions outputted by all base methods and bag of words. Here, we used all best parameters discovered in previous sections, including the random forest classifier. Note that the combination of these two set of features improves results compared with each single one separately. Although BoW individually presented better results in a few datasets it is not the best in all of them and alone, which suggests  that using both sets of features is the best option for 10SENT. In the next experiments, we always use this joint representation (BoW + BaseMethods) when me mention 10SENT.


\begin{table}[!htpb]
    \centering
    \small
    \begin{tabular}{|l|l|l|l|}
    \hline
    Dataset                & Bag of Words & BaseMethods & BoW + BaseMethods \\ \hline
    english\_dailabor      & 68.4        & 67.1       & 72.4             \\ \hline
    aisopos\_ntua          & 72.3        & 62.0       & 69.9             \\ \hline
    tweet\_semevaltest     & 58.3        & 62.8       & 65.2             \\ \hline
    sentistrength\_twitter & 58.8        & 59.1       & 61.2             \\ \hline
    sentistrength\_youtube & 56.6        & 56.1       & 58.7             \\ \hline
    sanders                & 61.5        & 54.1       & 56.4             \\ \hline
    sentistrength\_myspace & 50.2        & 52.3       & 52.2             \\ \hline
    sentistrength\_digg    & 45.4        & 50.1       & 50.6             \\ \hline
    nikolaos\_ted          & 51.3        & 45.9       & 49.0             \\ \hline
    debate                 & 57.1        & 45.9       & 47.1             \\ \hline
    sentistrength\_rw      & 48.3        & 48.5       & 45.5             \\ \hline
    sentistrength\_bbc     & 34.8        & 45.5       & 43.8             \\ \hline
    vader\_nyt             & 28.0        & 38.9       & 39.2             \\ \hline
    \end{tabular}
    \caption{Results of 10SENT using different set of features for  Random Forest}
    \label{tab:bow}
\end{table}

\subsection{Transfer Learning Analysis}

Finally, we evaluate whether it is possible to explore some ``easily available'' knowledge from an external source.
We do so by exploring datasets in which messages are labeled with ``emoticons'' by the systems' users themselves. To use an approach that transfers knowledge from one task to another, it is usually necessary to map characteristics from the source problem into the target one,  identifying similarities and differences. Next we detail how we transfer knowledge from existing emoticons in the datasets to the task of sentence-level sentiment analysis.




Emoticons are representations of an expression in a faced-look set of characters. They became very popular nowadays and even the Oxford English Dictionary has recently chosen an ``emoji'' as word of year (in 2015) due to its notable and massive use around the world.
In our case, they are used to give us an idea of feelings in the text, like happiness or sadness.

Previous works have demonstrated that such messages, though not available in large volumes, are very precise.
In other words, labeling with emoticons used by the final user indeed provide trustful information about the polarity of message.
Accordingly, in these experiments, we used the ``rules of thumb'' suggested in \citep{gonccalves2013comparing} to translate emoticons into polarities.

As one might expect, the fraction of messages containing emoticons is very low compared to the total number of messages.As we can see in Table \ref{tab:emoticoncoverage} emoticons appeared just in a very small amount of instances (observed in the coverage column). In spite of that, the accuracy of emoticons is often very precise to distinguish polarity of sentiment, reaching more than 90\% in ``nikolaos\_ted'' dataset. This is also in agreement with previous efforts~\citep{gonccalves2013comparing}.


Our ultimate goal here is to extract some information about the text of those messages to our classification step. For this, we incorporate into the training data these instances labeled with emoticons extracted from the respective datasets.

\begin{table}[!htpb]
    \centering
    \small
    \begin{tabular}{l|c|c|}
        \cline{2-3}
                                                     & Accuracy & Coverage \\ \hline
        \multicolumn{1}{|l|}{nikolaos\_ted}          & 0.919    & 0.014    \\ \hline
        \multicolumn{1}{|l|}{sentistrength\_myspace} & 0.800    & 0.091    \\ \hline
        \multicolumn{1}{|l|}{aisopos\_ntua}          & 0.787    & 0.526    \\ \hline
        \multicolumn{1}{|l|}{tweet\_semevaltest}     & 0.693    & 0.071    \\ \hline
        \multicolumn{1}{|l|}{english\_dailabor}      & 0.687    & 0.064    \\ \hline
        \multicolumn{1}{|l|}{sentistrength\_youtube} & 0.686    & 0.085    \\ \hline
        \multicolumn{1}{|l|}{sentistrength\_twitter} & 0.627    & 0.097    \\ \hline
        \multicolumn{1}{|l|}{sentistrength\_rw}      & 0.619    & 0.148    \\ \hline
        \multicolumn{1}{|l|}{sentistrength\_digg}    & 0.600    & 0.028    \\ \hline
        \multicolumn{1}{|l|}{sanders}                & 0.359    & 0.045    \\ \hline
        \multicolumn{1}{|l|}{debate}                 & 0.339    & 0.015    \\ \hline
        \multicolumn{1}{|l|}{sentistrength\_bbc}     & 0.173    & 0.006    \\ \hline
        \multicolumn{1}{|l|}{vader\_nyt}             & -        & -        \\ \hline
    \end{tabular}
    \caption{Accuracy and coverage of emoticons in training experiments for all datasets}
    \label{tab:emoticoncoverage}
\end{table}


To compare the effect of transfer learning from emoticons, we separated it in three different experiments: first with our traditional 10SENT; next we used just emoticon labels to create the training, without our majority voting predictions; then we combined these two to check the impact of emoticons in our method.
Results of this experiment can be seen in Table \ref{tab:transfer}. We can see that improvements of up to 6\% (e.g., in case of the sentistrength\_myspace dataset) can be obtained in terms of Macro F1, with no significant losses in most datasets and with no extra (labeling) cost. Thus, this approach represents an interesting opportunity to provide to the user some help in terms of labeling effort.

\begin{table}[!htpb]
    \centering
    \small
    \begin{tabular}{l|c|c|c|}
    \cline{2-4}
                                                 & 10Sent & Emoticons & 10Sent + Emoticons \\ \hline
    \multicolumn{1}{|l|}{english\_dailabor}      & 70.62  & 25.57     & 72.02              \\ \hline
    \multicolumn{1}{|l|}{aisopos\_ntua}          & 69.91  & 35.48     & 73.61              \\ \hline
    \multicolumn{1}{|l|}{tweet\_semevaltest}     & 64.78  & 18.06     & 65.13              \\ \hline
    \multicolumn{1}{|l|}{sentistrength\_twitter} & 62.17  & 22.93     & 62.87              \\ \hline
    \multicolumn{1}{|l|}{sentistrength\_youtube} & 57.06  & -         & 59.36              \\ \hline
    \multicolumn{1}{|l|}{sanders}                & 56.19  & 11.83     & 56.78              \\ \hline
    \multicolumn{1}{|l|}{sentistrength\_digg}    & 51.91  & -         & 52.22              \\ \hline
    \multicolumn{1}{|l|}{sentistrength\_myspace} & 50.22  & -         & 53.20              \\ \hline
    \multicolumn{1}{|l|}{nikolaos\_ted}          & 47.97  & -         & 48.97              \\ \hline
    \multicolumn{1}{|l|}{debate}                 & 47.18  & -         & 47.37              \\ \hline
    \multicolumn{1}{|l|}{sentistrength\_rw}      & 47.15  & -         & 45.25              \\ \hline
    \multicolumn{1}{|l|}{sentistrength\_bbc}     & 43.76  & -         & 43.18              \\ \hline
    \multicolumn{1}{|l|}{vader\_nyt}             & 39.81  & -         & 39.01              \\ \hline
    \end{tabular}
    \caption{Macro-F1 results for experiments on 10SENT using Transfer Learning}
    \label{tab:transfer}
\end{table}




 \section{Comparative Results}	

We now turn our attention to the comparison between 10SENT, the ``strongest'' baseline (Majority voting) and the base methods. We should point out that in these comparisons, a mention to 10SENT corresponds to the results obtained with the best unsupervised configuration found in the previous analyses, in other words, the original 10SENT representation (methods' decisions) along with the Bag Of Words and the transfer learning.

\begin{figure}[!htpb]
  	
  	\centering
{
		\includegraphics[width=\textwidth]{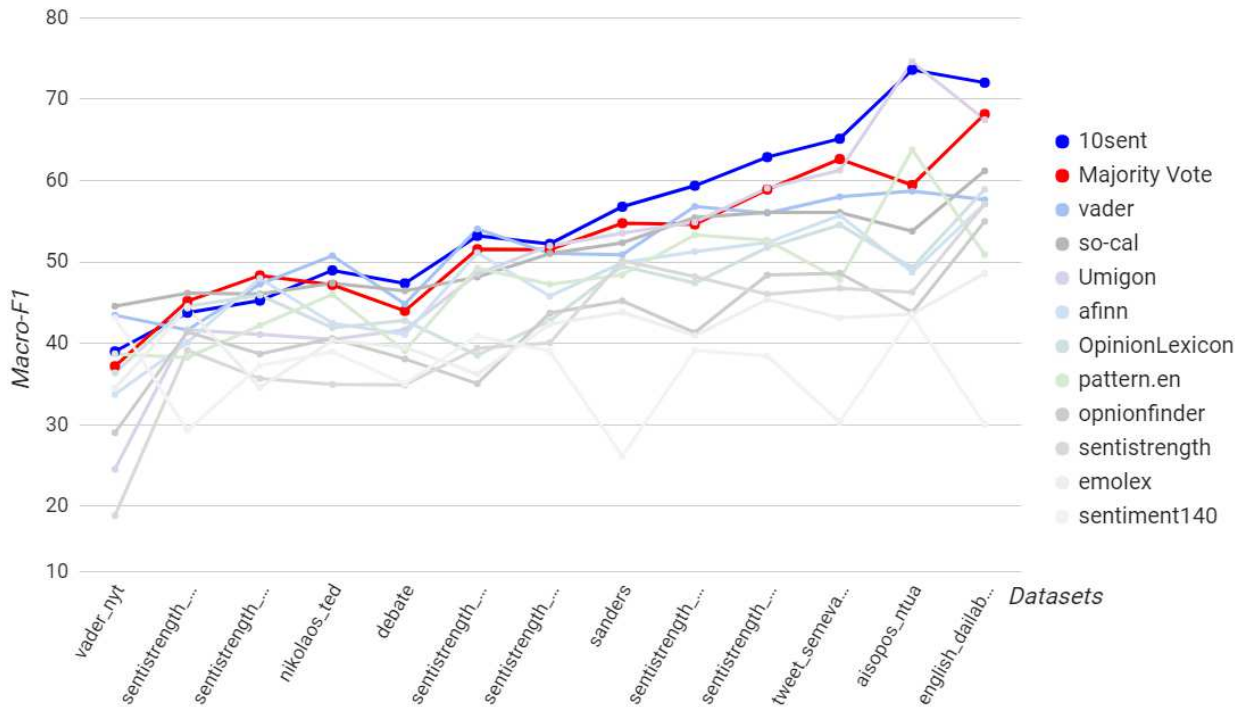}		   	 }
 	 \caption{Macro-F1 results of 10SENT compared with each individual base method for all datasets}
	\label{fig:macroF1}

\end{figure}

We can observe in Figure \ref{fig:macroF1} that our method has a higher Macro-F1,  above the baselines, in most datasets. In fact, 10SENT is the best method in 7 out of 13 datasets and it is close to the top of the rank in several others. This is also reflected in the Mean Rank, shown in Table \ref{table:rank}, confirming that 10SENT is the overall winner across all tested datasets.

\begin{table}[!htpb]
    \centering
    \small
    \begin{tabular}{|l|l|l|l|}
        \hline
        \textbf{METHOD}      & \textbf{MEAN RANK} & \textbf{POS} & \textbf{DEVIATION} \\ \hline
        10SENT               & 2.154              & 1            & 1.457              \\ \hline
        Majority Voting      & 3.154              & 2            & 1.350              \\ \hline
        Vader                & 3.692              & 3            & 1.814              \\ \hline
        SO-CAL               & 3.769              & 4            & 1.717              \\ \hline
        Umigon               & 4.923              & 5            & 2.921              \\ \hline
        Afinn                & 6.615              & 6            & 1.820              \\ \hline
        OpinionLexicon       & 6.923              & 7            & 1.900              \\ \hline
        pattern.en           & 7.000              & 8            & 2.287              \\ \hline
        OpinionFinder        & 9.308              & 9            & 1.136              \\ \hline
        Sentistrength        & 9.846              & 10           & 1.747              \\ \hline
        Emolex               & 9.923              & 11           & 2.055              \\ \hline
        Sentiment140 Lexicon & 10.692             & 12           & 2.493              \\ \hline
    \end{tabular}
    \caption{Mean Rank of methods for all datasets}
    \label{table:rank}
\end{table}	

In fact, 10SENT can be considered as the most \textit{stable method} as it produces the best (or close to the best) results in most datasets in different domains and applications. In other words, by using our proposed method, one can almost always guarantee top-notch results, at no extra cost, and without the need to discover the best method for a given context/dataset/domain.


%

\subsection{UpperBound Comparison}

For analysis purposes, we perform a comparison of 10SENT with some ``uppperbound'' baselines which use some type of privileged information, most notably the real label of the instances in the training set, an information  not available to us. The idea here it to understand how far our proposed unsupervised approach is from the ones that exploit such information as well as to understand the limits to what we can achieve with an unsupervised solution.

The first ``upperbound'' baseline is a fully supervised approach which uses all the labeled information available in the training data. As  normally done in fully supervised approaches, the parameters of the RF algorithm are determined using a validation set.

The second baseline is an \textit{Exhaustive Weighted Majority Voting} method that uses the real labels of messages of the datasets to find the best possible linear combination of weights for each base method. Differently from the Majority Voting baseline, in which all methods have the same weight, in this approach, each individual base method has a different weight, so that the influence of each one in the final classification is different.

The weights for each method are found by means of an exhaustive search in each (training portion of the) dataset. That is, for each dataset we found the “close-to-ideal'' weights that would lead to the best possible result when combining the exploited base methods. Then, for each method, a weight was associated with its output and, finally, the class with the highest weight was marked as  the resulting label of each instance.  This search was performed in exhaustive mode, i.e., we evaluate every possible combination, seeking to maximize the Macro-F1 in each dataset.
During the experiments, we limited the search to five different weights in the range $[0-1]$: $ W = \{0, 0.25,  0.5, 0,75, 1\}$) to estimate ``close-to-best''  results, while maintaining feasible computational costs.


Table \ref{table:weights} shows the average weights and corresponding standard deviation for each method in some datasets, but for all the results are similar. We can see that most methods have different behaviors in different datasets (implied by the large deviations). In other words, the same method may have a huge variance in effectiveness in different datasets, which precludes the use of a single unique method for all cases. Despite this, we can observe that some methods have clearly a higher average than others even with this high deviation.

\begin{table}[!htpb]
\centering
\small
\begin{tabular}{l|l|l|l|l|l|}
\cline{2-6}
\multicolumn{1}{c|}{}                & \multicolumn{5}{c|}{Weights}                                       \\ \cline{2-6}
                                     & pattern.en & sentiment140 & emolex & opinionfinder & sentistrength \\ \hline
\multicolumn{1}{|l|}{Avg. Weight}    & 0.28       & 0.37         & 0.26   & 0.40          & 0.85          \\ \hline
\multicolumn{1}{|l|}{Std. Deviation} & 0.25       & 0.28         & 0.29   & 0.31          & 0.28          \\ \hline
\end{tabular}

\caption{Average and deviation for weights found during Exhaustive Weighted Vote step}
\label{table:weights}

\end{table}
\normalsize

Finally, the third ``upperbound'' baseline is the best single base method in each dataset. Since the base methods are unsupervised ``off-the-shelf'' ones, we determine the best method for each dataset also using the labels in the training sets. It is also an ``upperbound'' because the best method cannot be determined, in advance, without supervision, i.e., a training set.

\subsubsection{Upperbound Results}

The results of those upperbounds are shown in Table \ref{table:upperbounds}. For comparative purposes we also included in this table the results of the unsupervised Majority Voting.
As before, all results correspond to the average performance in the 5 test sets of the folded cross-validation procedure using 10SENT with its best configuration including Bag of Words and Transfer Learning.

Values marked with ``\textbf{*}'' in this table indicate that the difference was not statistically significant when compared to  10SENT in a paired-test with 95\% confidence.
Results reported with ``\textsuperscript{$\triangle$}'' are those statistically better than those of 10SENT.
On the other hand, our method demonstrated to be statistically superior to the ones whose values are marked with ``\textsuperscript{$\nabla$}''.

\begin{table}[!htpb]
  \centering
  \small
  \begin{tabular}{|l|l|l|l|l|l|}
  \hline
                         & \multicolumn{1}{c|}{Fully Supervised} & \multicolumn{1}{c|}{\begin{tabular}[c]{@{}c@{}}Exhaustive Weighted \\ Majority Voting\end{tabular}} & \multicolumn{1}{c|}{\begin{tabular}[c]{@{}c@{}}Best\\ Individual\end{tabular}} & \multicolumn{1}{c|}{Majority Voting} & \multicolumn{1}{c|}{10SENT} \\ \hline
  aisopos\_ntua          & 76.64\textsuperscript{$\triangle$}                                 & 75.8\textsuperscript{$\triangle$}                                                                                                & 74.58*                                                                          & 59.45\textsuperscript{$ \nabla$}                                 & 73.61                       \\ \hline
  english\_dailabor      & 75.63\textsuperscript{$\triangle$}                                & 71.9*                                                                                                & 67.47\textsuperscript{$\nabla$}                                                                          & 68.16\textsuperscript{$\nabla$}                                & 72.02                       \\ \hline
  tweet\_semevaltest     & 66.77*                                 & 65.5*                                                                                                & 61.27\textsuperscript{$\nabla$}                                                                          & 62.64\textsuperscript{$\nabla$}                                & 65.13                       \\ \hline
  sentistrength\_twitter & 66.14\textsuperscript{$\triangle$}                                 & 62.9*                                                                                                & 59.05\textsuperscript{$\nabla$}                                                                          & 58.89\textsuperscript{$\nabla$}                               & 62.87                       \\ \hline
  sentistrength\_youtube & 61.77\textsuperscript{$\triangle$}                                 & 60.6*                                                                                                & 56.81\textsuperscript{$\nabla$}                                                                          & 54.60\textsuperscript{$\nabla$}                               & 59.36                       \\ \hline
  sanders                & 62.76\textsuperscript{$\triangle$}                                 & 58.0\textsuperscript{$\triangle$}                                                                                                  & 53.52*                                                                         & 54.75*                               & 56.78                       \\ \hline
  sentistrength\_myspace & 57.47\textsuperscript{$\triangle$}                                & 57.8\textsuperscript{$\triangle$}                                                                                                & 54.05*                                                                          & 51.56*                               & 53.20                       \\ \hline
  sentistrength\_digg    & 59.52\textsuperscript{$\triangle$}                                 & 57.3\textsuperscript{$\triangle$}                                                                                                & 51.98*                                                                         & 51.50*                               & 52.22                       \\ \hline
  nikolaos\_ted          & 57.43\textsuperscript{$\triangle$}                                 & 56.1\textsuperscript{$\triangle$}                                                                                                & 50.76\textsuperscript{$\triangle$}                                                                         & 47.17*                               & 48.97                       \\ \hline
  debate                 & 58.75\textsuperscript{$\triangle$}                                 & 49.1\textsuperscript{$\triangle$}                                                                                                & 46.45*                                                                         & 43.99\textsuperscript{$\nabla$}                                & 47.37                       \\ \hline
  sentistrength\_rw      & 53.52\textsuperscript{$\triangle$}                                 & 52.2\textsuperscript{$\triangle$}                                                                                                & 47.97*                                                                         & 48.34\textsuperscript{$\triangle$}                               & 45.25                       \\ \hline
  sentistrength\_bbc     & 44.00*                                & 51.8\textsuperscript{$\triangle$}                                                                                                & 46.17*                                                                         & 45.19*                               & 43.18                       \\ \hline
  vader\_nyt             & 46.87\textsuperscript{$\triangle$}                                 & 51.9\textsuperscript{$\triangle$}                                                                                                & 44.56\textsuperscript{$\triangle$}                                                                          & 37.19\textsuperscript{$\nabla$}                                & 39.01                       \\ \hline
  \end{tabular}
	\caption{Results in terms of Macro-F1 comparing 10SENT with all other evaluation methods (``*'' indicates values that the difference was not statistically significant compared to the 10SENT; ``\textsuperscript{$\nabla$}'' are values that 10SENT wins and ``\textsuperscript{$\triangle$}'' are the values statistically superior to the 10SENT result)}
	\label{table:upperbounds}
\end{table}

As highlighted before, 10SENT is tied or better than the traditional majority voting in most datasets, being statistically superior in seven out of 12 cases, tying in other 5 and losing only in one dataset (sentistrength\_rw). Gains can achieve up to 23.8\% against this baseline.
When compared to the best individual method in each dataset, 10SENT wins (4 cases) or ties (7 cases)  in 11 out of 13 cases, a strong result. This shows that 10SENT is a good and consistent choice among all available options, independently of which dataset is used.

When compared to supervised Exhaustive Weighted Majority Voting, a first observation is that, as expected, it is always superior to the simple Majority Voting. Although we cannot beat this ``upperbound'' baseline, we tie with it in 4 datasets (sentistrenth\_youtube, sentistrength\_twitter, tweet\_semevaltest, english\_dailabor) and get close results in others such as aisopos\_ntua, sanders and debate. This with no cost at all in terms of labeling effort.

Regarding the strongest upperbound baseline -- Fully Supervised --, an interesting observation to make is that in some datasets its results get very close to those of the Exhaustive Weighted Majority Voting, even loosing to it in two (sentistrength\_bbc, vader\_nyt). This is a surprising result, meaning that the combination of both strategies is also an interesting venue to pursue in the future. When comparing this baseline to 10SENT, as expected, we can also not beat it, but can tie with it in two datasets and get close results in others, mainly in those cases in which our method was a good competitor against Exhaustive Weighted Majority Voting. We consider these very strong results.


For a deeper understanding of the results, Table \ref{tab:analysis} shows the set size of 10SENT used to train the classifier before and after the bootstrapping  step (lines 9-11 of Algorithm 1).
As we can see, the majority voting heuristics selects a relative large amount of training data from the original datasets. This may explain some of the good results obtained in our experiments, since the classifiers have a reasonable amount of data to be trained with.

\begin{table}[!htpb]
    \centering
    \small
    \begin{tabular}{l|l|l|l|l|l|l|}
    \cline{2-7}
    \multirow{3}{*}{}                                                                       & \multicolumn{6}{c|}{10SENT}                                                                                                                                                       \\ \cline{2-7}
                                                                                            & \multicolumn{3}{c|}{Majority Voting}                                                      & \multicolumn{3}{c|}{Bootstrapping}                                                      \\ \cline{2-7}
                                                                                            & \multicolumn{1}{c|}{Set Size} & \multicolumn{1}{c|}{Accuracy} & \multicolumn{1}{c|}{F1} & \multicolumn{1}{c|}{Set Size} & \multicolumn{1}{c|}{Accuracy} & \multicolumn{1}{c|}{F1} \\ \hline
    \multicolumn{1}{|l|}{english\_dailabor}                                                 & 1999                          & 0.858                         & 70.61                   & 2165                          & 0.826                         & 70.62                   \\ \hline
    \multicolumn{1}{|l|}{aisopos\_ntua}                                                     & 215                           & 0.762                         & 71.67                   & 238                           & 0.734                         & 69.91                   \\ \hline
    \multicolumn{1}{|l|}{tweet\_semevaltest}                                                & 2781                          & 0.796                         & 65.04                   & 3139                          & 0.757                         & 64.78                   \\ \hline
    \multicolumn{1}{|l|}{\begin{tabular}[c]{@{}l@{}}sentistrength\_\\ twitter\end{tabular}} & 2042                          & 0.706                         & 63.06                   & 2238                          & 0.665                         & 62.17                   \\ \hline
    \multicolumn{1}{|l|}{\begin{tabular}[c]{@{}l@{}}sentistrength\_\\ youtube\end{tabular}} & 1688                          & 0.660                         & 58.37                   & 1837                          & 0.645                         & 57.06                   \\ \hline
    \multicolumn{1}{|l|}{sanders}                                                           & 1760                          & 0.762                         & 55.94                   & 1929                          & 0.758                         & 56.19                   \\ \hline
    \multicolumn{1}{|l|}{sentistrength\_digg}                                               & 474                           & 0.630                         & 49.32                   & 519                           & 0.615                         & 51.91                   \\ \hline
    \multicolumn{1}{|l|}{\begin{tabular}[c]{@{}l@{}}sentistrength\_\\ myspace\end{tabular}} & 488                           & 0.641                         & 49.34                   & 535                           & 0.645                         & 50.22                   \\ \hline
    \multicolumn{1}{|l|}{debate}                                                            & 1422                          & 0.508                         & 47.30                   & 1620                          & 0.530                         & 47.97                   \\ \hline
    \multicolumn{1}{|l|}{nikolaos\_ted}                                                     & 329                           & 0.606                         & 47.11                   & 370                           & 0.556                         & 47.18                   \\ \hline
    \multicolumn{1}{|l|}{sentistrength\_rw}                                                 & 417                           & 0.618                         & 43.12                   & 471                           & 0.601                         & 47.15                   \\ \hline
    \multicolumn{1}{|l|}{sentistrength\_bbc}                                                & 376                           & 0.687                         & 37.91                   & 418                           & 0.661                         & 43.76                   \\ \hline
    \multicolumn{1}{|l|}{vader\_nyt}                                                        & 2222                          & 0.363                         & 40.44                   & 2563                          & 0.366                         & 39.81                   \\ \hline
    \end{tabular}
    \caption{Set size, ``noise''(indicated by accuracy) and Macro-F1 values to 10SENT training sets without bootstrapping and including bootstrapping step}
    \label{tab:analysis}
\end{table}

However, this is only part of the story. One question that remains to be answered is: ``What is the \textbf{quality} of the automatically labeled training set''. We can answer this question by looking at the columns ``Accuracy'' in the table. This metric calculates the proportion of correctly assigned labels in the training sets when compared to the ``real labels''. For a considerable number of datasets, the accuracy is relatively high, between 0.6-0.8. In fact,  the cases in which 10SENT gets closer to the fully supervised method correspond to those in which the accuracy in the training is higher. We can also see that after the bootstrapping, in general the accuracy in the training drops a bit, which is natural since the heuristics is not perfect, but this is compensated by the increase in training size, resulting in a learned model that generalizes better.

Finally, we can see that the absolute results of the best overall method in each dataset are still not very high (maximum of 76\%), which shows the difficulty of the sentiment analysis task and that there is a lot of room for improvements.


\section{Conclusions}

We presented a novel unsupervised approach for sentiment analysis on sentence-level derived from the combination of several existing ``off-the-shelf'' sentiment analysis methods. Our solution was thoroughly tested in a wide and diversified environment. We cover a vast amount of methods and labeled datasets from different domains. The key advantage of 10SENT is that it fixes  one major issue in this field -- the variability of the methods across domains and datasets. Our experimental results show that our self-learning approach has the lowest prediction performance variability due to its ability to slightly adapt  to different contexts.  This is a crucial issue in an area in which researchers are mostly interested in using an ``off-the-shelf'' method to different contexts. Our approach is also easily expandable to include any new developed unsupervised method. Our experimental results also show that 10SENT achieves good effectiveness compared to our baselines. 10SENT was superior to all existing individual methods and also obtained better results than the traditional majority voting, with gains of up to 17.5\%. In an upperbound comparison, we saw that 10SENT can get close to the best supervised results. Finally, our analysis on transfer learning shows us the possibility of adapting the method to include more strategies that can lead to better results. As future work, we intend to better explore weights as well choosing other setups for different scenarios. Additionally, we will explore other syntactic and semantic aspects of the text of the messages to improve results. We also plan to release our codes and datasets to the research community and deploy our method as part of known sentiment analysis benchmark systems~\citep{araujo2016@icwsm}.



\bibliographystyle{apalike}

\end{document}